\documentclass[letterpaper, 10 pt, journal, twoside]{IEEEtran}
\usepackage{amsmath}
\usepackage{amssymb}
\usepackage{graphicx}
\usepackage[hidelinks]{hyperref}
\usepackage{float} 
\usepackage[ruled,noend]{algorithm2e} 

\usepackage{booktabs}
\usepackage{siunitx}
\usepackage{enumerate}
\usepackage{xcolor}
\usepackage{wrapfig}
\usepackage{hyperref}
\usepackage{caption} 

\definecolor{revblue}{RGB}{0, 0, 200}

\begin{document}
%
\title{ZeroBot: Learning from Scratch in Minutes\\with Generative Real2Sim}
%
%
%

\author{Ivan Kapelyukh$^{1,2}$, Xiaohan Zhang$^2$, Stephen James$^1$, Laura Herlant$^2$, Edward Johns$^1$%
\thanks{Received 19 August 2025. Accepted 7 January 2026. Date of publication
9 February 2026. Date of current version 27 February 2026.}
\thanks{This article was recommended for publication by Associate Editor Y. Li and Editor M. Vincze upon evaluation of the reviewers’ comments.}
\thanks{$^1$Ivan Kapelyukh, Stephen James, and Edward Johns are at Imperial College London. Ivan Kapelyukh is with the Robot Learning Lab and the Dyson Robotics Lab. Stephen James is with the Safe Whole-body Intelligent Robotics Lab (SWIRL). Edward Johns is with the Robot Learning Lab.}
\thanks{$^2$Xiaohan Zhang and Laura Herlant are at the Robotics and AI Institute. Part of this work was conducted during Ivan Kapelyukh's internship there.}
\thanks{Digital Object Identifier 10.1109/LRA.2026.3662595}
}
%
%

\markboth{IEEE Robotics and Automation Letters. Preprint Version. Accepted January, 2026}
{Kapelyukh \MakeLowercase{\textit{et al.}}: ZEROBOT: LEARNING FROM SCRATCH IN MINUTES WITH GENERATIVE REAL2SIM} 

%


\maketitle

\begin{abstract}
We present ZeroBot, a real2sim framework for learning a robot manipulation task from scratch in minutes under challenging conditions: zero human demonstrations, zero policy pre-training, and zero known object models. Given only a single view of an object and a goal pose for that object, ZeroBot uses image-to-3D generative models to obtain a complete object mesh, which is used in simulation for large-scale parallel reinforcement learning. To accelerate training, we introduce an action space which leverages the generated geometry and learned value function to sample states involving robot-object contact. When evaluated on real-world tasks including grasping, pushing, articulated object interaction, and multi-stage manipulation, ZeroBot achieves an 87\% success rate with an average training time of 119 seconds. These results show the value of using image-to-3D models in a real2sim framework for rapid, autonomous robot learning. Project page: \textcolor{blue}{\href{https://zerobot-rl.github.io}{zerobot-rl.github.io}}
\end{abstract}


%
\IEEEpeerreviewmaketitle

%
%
%
%


\section{Introduction}

\IEEEPARstart{W}{e} propose a new and challenging problem setting: a robot must manipulate a novel object into a given goal pose, with zero demonstrations, zero policy pre-training, and zero object models known in advance. The method should learn from scratch in minutes, and should have the versatility to learn tasks involving grasping, pushing, rolling, multi-stage interaction, or generalization to new initial poses. This is an important research direction, as it would enable robots to learn new tasks with minimal human effort. We present ZeroBot, the first method to address this problem setting.

To perform specific tasks such as pick-and-place, classical pipelines use grasp sampling and kinematic motion planning \cite{kpam,llm-grop}. However, these pipelines often lack the versatility needed to also learn behaviors such as rolling or tasks with multiple contact modes. Dynamics-based trajectory optimization \cite{drop,global-planning} can be used instead, but often requires accurate dynamics models. The desire for generalization and more robust closed-loop behavior motivates policy learning.

Reinforcement learning (RL) is a popular method for learning a policy autonomously \cite{rl-survey}. However, real-world RL is limited by the time taken to collect real-world data, and often requires human supervision for safety and environment resetting. Training in simulation is a promising alternative as it scales well with available compute \cite{isaac-gym,dextreme}. The challenge becomes building a simulation from which policies transfer to the real world. One approach is to generate a large dataset of scenes for simulation training, hoping that this covers the situations that the robot may encounter at deployment \cite{manipgen,cabinet}, but generating diverse, realistic scenes is challenging.

\begin{figure}[t]
    \centering
    \includegraphics[width=1.0\linewidth]{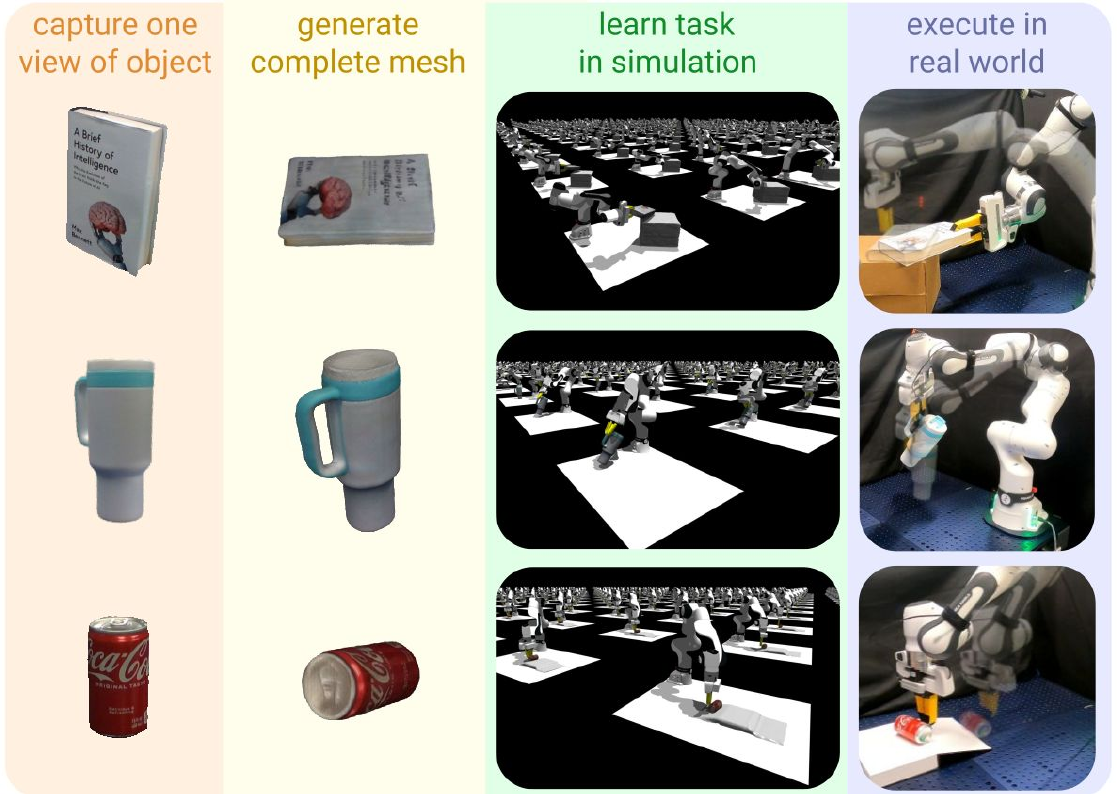}
    \caption{From a single view of an object, our ZeroBot framework generates a complete object mesh using an image-to-3D model, which is used in simulation for massively parallel reinforcement learning with an efficient action space. The policy is learned in minutes, with no human demonstrations or policy pre-training, and deployed zero-shot on the real robot.}
    \label{fig:teaser}
\end{figure}

Instead, real2sim methods build a simulation of the scene observed during deployment. Many existing real2sim methods either assume known object models \cite{closing-loop}, or require a human to manually scan the objects \cite{rialto}. In this work we show how the visual prior of image-to-3D generative models can enable robots to generate a complete object mesh for real2sim from only a single view (Figure \ref{fig:teaser}). We adopt this approach as it eliminates the need for object scanning, which can be time-consuming, often requires human involvement, and is inherently unable to reconstruct occluded surfaces, thus motivating our investigation into 3D priors.

Once the simulation has been built, we use massively parallel RL to learn a policy that can manipulate an object into a given goal pose. We develop an action space which leverages the generated object geometry to sample states where the robot is in contact with the object, significantly accelerating learning. At deployment, the learned value function from RL is used to select a high-value contact location. After moving into contact, the robot then executes actions using its policy. This leads to a system for rapidly learning to manipulate a new object with minimal human effort. Our contributions include:

\begin{enumerate}
    \item
    A new problem setting which requires learning a manipulation task with zero human demonstrations, zero policy pre-training, and zero prior object models. We demonstrate that this is possible with our ZeroBot framework, designed for learning from scratch in minutes.

    \item
    A real2sim-and-sim2real system which builds a simulation of a scene from a single view in seconds using pre-trained vision foundation models. We show how the generated mesh can be used for learning in simulation with an efficient action space, as well as during deployment for collision-free motion planning and object tracking, which enables zero-shot sim2real transfer.

    \item
    A novel action space designed for rapid learning, which leverages the generated geometry to select high-value contact states. This enables efficient exploration, removes the need for reward engineering, and can learn tasks such as grasping, pushing, articulated manipulation, and multi-stage behavior. We show empirically that this is the critical factor which enables learning policies in minutes.

    \item
    Experiments evaluating 3D generative models for real2sim. We demonstrate that mesh generation leads to higher sim2real success rates than mesh retrieval, or using partial meshes. We further study the performance gap between single-view generation and ground-truth mesh scanning, and demonstrate that sim2real transfer is possible even when task-critical geometry is unobserved.
    
\end{enumerate}
To the best of our knowledge, this is the first work which addresses this challenging problem setting, and shows how image-to-3D models can be used to learn manipulation tasks autonomously from scratch in minutes.

\section{Related Work}
\label{s:related-work}
\textbf{Reinforcement learning} can be used to learn a policy autonomously through exploration \cite{rl-survey}. Real-world RL has been made more time-efficient through algorithmic improvements in data efficiency \cite{mpo}, careful system design \cite{serl}, and efficient action spaces, for example using iterative discretization \cite{crl} or spatial grounding \cite{hacman, hacmanpp}. Prior work on efficient action spaces typically focuses on generalization to novel objects and can take several days to train \cite{hacmanpp}. We develop an action space which is tailored to our problem setting of learning from scratch in minutes, by leveraging the generated object geometry. While real-world RL is bottlenecked by the rate of real-world data collection, simulation enables learning from many robots in parallel, scaling better with compute.

\textbf{Real2sim} methods build a simulation of the scene in front of the robot. Some methods infer the physics properties of objects, such as mass, friction, and articulation \cite{closing-loop,asid,contact-dynamics,real2sim2real-casting,urdformer}. This is complementary to our work, which generates object geometry from a single visual observation. For visual real2sim, NeRF \cite{rialto} or Gaussian Splatting can be used \cite{embodied-gaussians}. Crucially, these methods typically do not use a visual prior to generate complete object geometry, instead requiring a human to manually scan objects, or set up multiple cameras. We show that recent advances in image-to-3D models allow a complete object mesh to be generated by a robot from a single view, making real2sim faster and more autonomous. We mitigate the visual sim2real gap \cite{sim2sim} by deploying a state-based policy at test time, using the generated mesh for tracking with Foundation Pose \cite{foundation-pose}.

\textbf{Image-to-3D generative models} build a 3D mesh from partial observations of an object \cite{instantmesh}. Shape completion has been used in robotics to aid collision detection \cite{cabinet}. However, 3D training data is not available at scale. A recent breakthrough has been the use of 2D image diffusion models as a powerful, web-scale visual prior for mesh generation, greatly enhancing generalization \cite{instantmesh}. These image-to-3D models have been used in robotics to aid grasp sampling with a pre-trained grasp network \cite{scenecomplete}. We study training an RL policy from scratch, without pre-training. Other methods use them to reconstruct videos of human demonstrations \cite{video2policy,phystwin}. They have also been used to generate assets for large-scale simulation training \cite{robogen,gen2sim,robotwin}. However, these methods primarily focus on offline data generation. For example, Gen2Sim \cite{gen2sim} uses score-distillation sampling, an offline method typically requiring significant optimization time, while RoboTwin \cite{robotwin} uses human demonstrations and functional part annotations. Our framework generates a mesh of an object in seconds and autonomously learns to manipulate it from scratch in minutes, with zero demonstrations.

\section{Method}
\subsection{Problem Setting and Assumptions}\label{ss:problem-setting}

Given a single RGB-D image $\mathcal{I}$ of an object, and a goal pose $\mathcal{T}_{OG}$ $\in$ SE(3) for that object, the robot must learn within minutes to manipulate the object into that goal pose. The goal pose is relative to the object's initial pose, and is defined by the user. This is sufficient to specify many everyday tasks \cite{rearrangement,cliport}. We do not assume access to any human demonstrations, policy pre-training, or known object models. In this work, we primarily focus on tasks where the robot interacts with a single, rigid object. We show that our framework can be extended to articulated objects in Section \ref{ss:artic-exp}. We assume that the provided camera view can see the object clearly, as the image-to-3D model which we use \cite{instantmesh} is trained on upright, occlusion-free views. We also assume a static scene background (such as shelves), which the robot scans automatically using its wrist camera before the dynamic object is placed into the scene. Image-to-3D models are rapidly improving, and future work can easily swap new models into our modular framework for scene-level reconstruction and multi-object manipulation, enabling more dynamic tasks. 

\begin{figure*}[ht]
    \centering
    \includegraphics[width=0.85\textwidth]{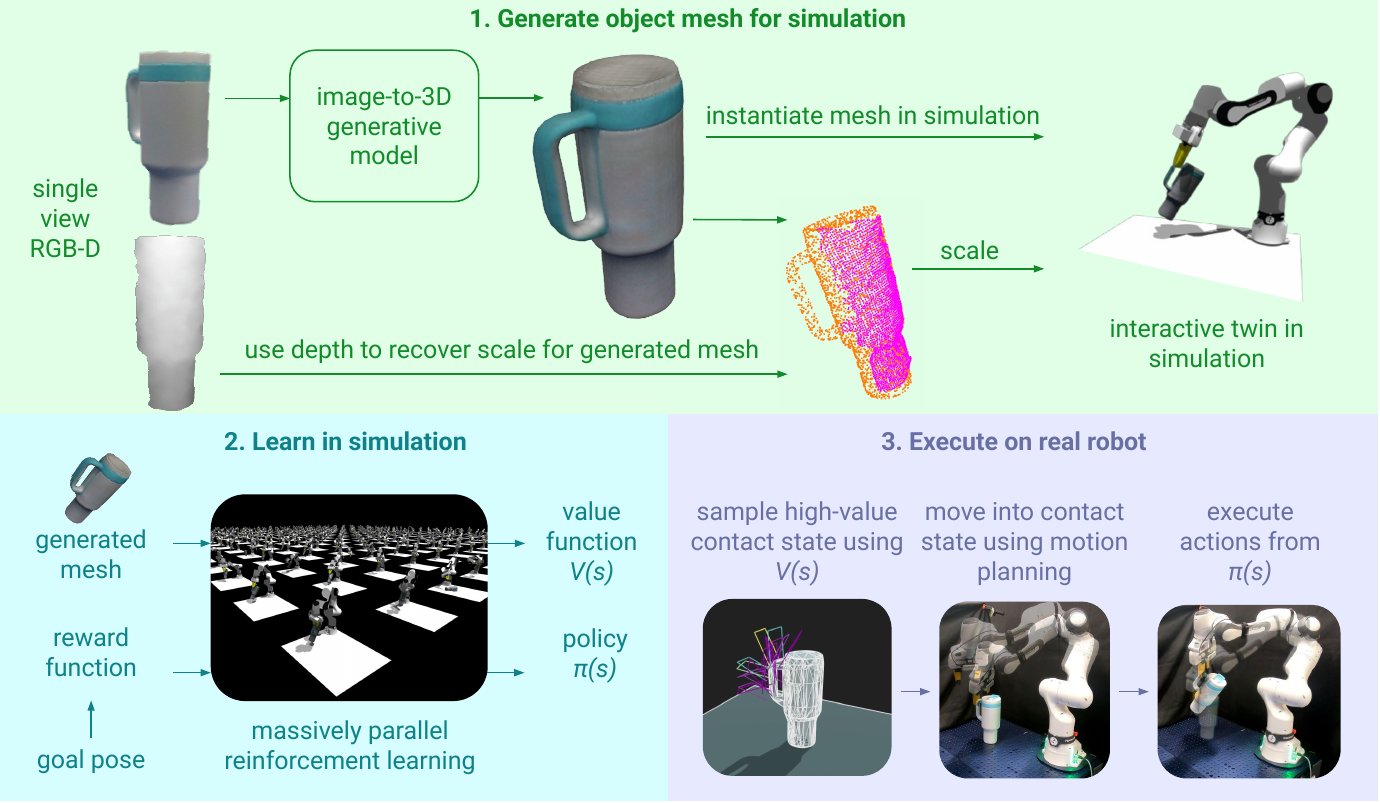}
    \caption{The three stages of our ZeroBot framework. First, the robot observes an object from a single view. It generates an object mesh from the RGB image using an image-to-3D model, and aligns it with the depth image to recover scale. Second, the generated mesh is used for simulation. The task is specified by providing a goal pose, which defines a dense reward function. Massively parallel RL is used to learn a policy which can manipulate the object into that goal pose. To accelerate training, we use a contact action space which selects a high-value contact state using the learned value function. During deployment, the robot moves into this state using motion planning. Once in contact, it executes actions from its policy.}
    \label{fig:pipeline}
    \vspace{-0.3cm}
\end{figure*}

To address this problem, we develop a real2sim framework shown in Figure \ref{fig:pipeline}. We now describe how the real2sim is performed (Section \ref{ss:build-sim}), and how a policy is learned rapidly in simulation using our action space, deploying zero-shot to the real robot (Section \ref{ss:sim-training}).

\subsection{Building a Simulation Using Generative Priors}\label{ss:build-sim}

The objective of the first stage is to generate an object mesh from a single view for use in simulation. The robot captures an RGB-D image using its wrist camera. In our experiments, the user specifies the target object by clicking on it to produce a segmentation mask using SAM 2 \cite{sam2}. In future work, this can be specified with language for convenience \cite{lseg}. The masked RGB image is used as input to the InstantMesh \cite{instantmesh} image-to-3D pipeline. First, a diffusion model generates novel views of the object, then a transformer predicts a 3D feature map, from which a mesh is extracted. This entire InstantMesh pipeline runs in 10 seconds on one desktop GPU. However, InstantMesh was not trained to predict metric scale. To recover the scale, we align the generated mesh with the partial pointcloud from the depth image from that same view. We use batched ICP \cite{icp} to optimize many randomly sampled alignments and then select the scale from the highest-fitness solution. The result is a colored mesh with metric scale.

Training a pixels-to-actions policy is infeasible in this problem setting, as we must train from scratch in minutes. Since we have a generated object mesh, we can use that for pose tracking, so that we can rapidly train and deploy a state-based policy. We use Foundation Pose \cite{foundation-pose}, which takes as input our generated mesh, and an RGB-D image, to predict the object's pose. We use Foundation Pose to infer the initial object pose for real2sim, and also during execution to provide the latest object pose as an input to our policy. Foundation Pose is trained automatically on over a million images of synthetic scenes. Integrating this into our framework allows our method to learn to manipulate a novel object in minutes. In future work, the state-based policy can be distilled \cite{manipgen} into a pixels-to-actions policy, if more training time is available.

\subsection{Simulation Training and Execution with Contact Actions}\label{ss:sim-training}

\textbf{Simulation training overview}. We train a policy $\pi(s)$ and value function $V(s)$ in Isaac Gym \cite{isaac-gym} using PPO \cite{ppo}, due to its time efficiency in the parallel simulation setting. We use the same simple, dense reward function of goal flow~\cite{hacmanpp} for all tasks: for each point on the object, we construct a goal flow vector pointing to where that point would be if the object were in its goal pose. Minimizing the goal flow norm would maximize the reward. This avoids the need for a hyperparameter to weigh between position and orientation error. The object's initial pose in simulation matches the real pose. It is dropped from a low height to avoid collisions with the background. We also randomize the initial pose if the experiment requires learning a policy which generalizes to novel poses. Although we \textit{do} include an experiment on generalization, it is not the focus of this paper: the robot should learn to complete the \textit{specific instance} of the given task from scratch autonomously in minutes. The state vector $s$ (input to the value function and policy) includes: the object position and quaternion, the gripper's left and right finger positions, the gripper quaternion, and a binary gripper open/closed indicator.

\begin{figure*}[ht]
    \centering
    \includegraphics[width=0.9\linewidth]{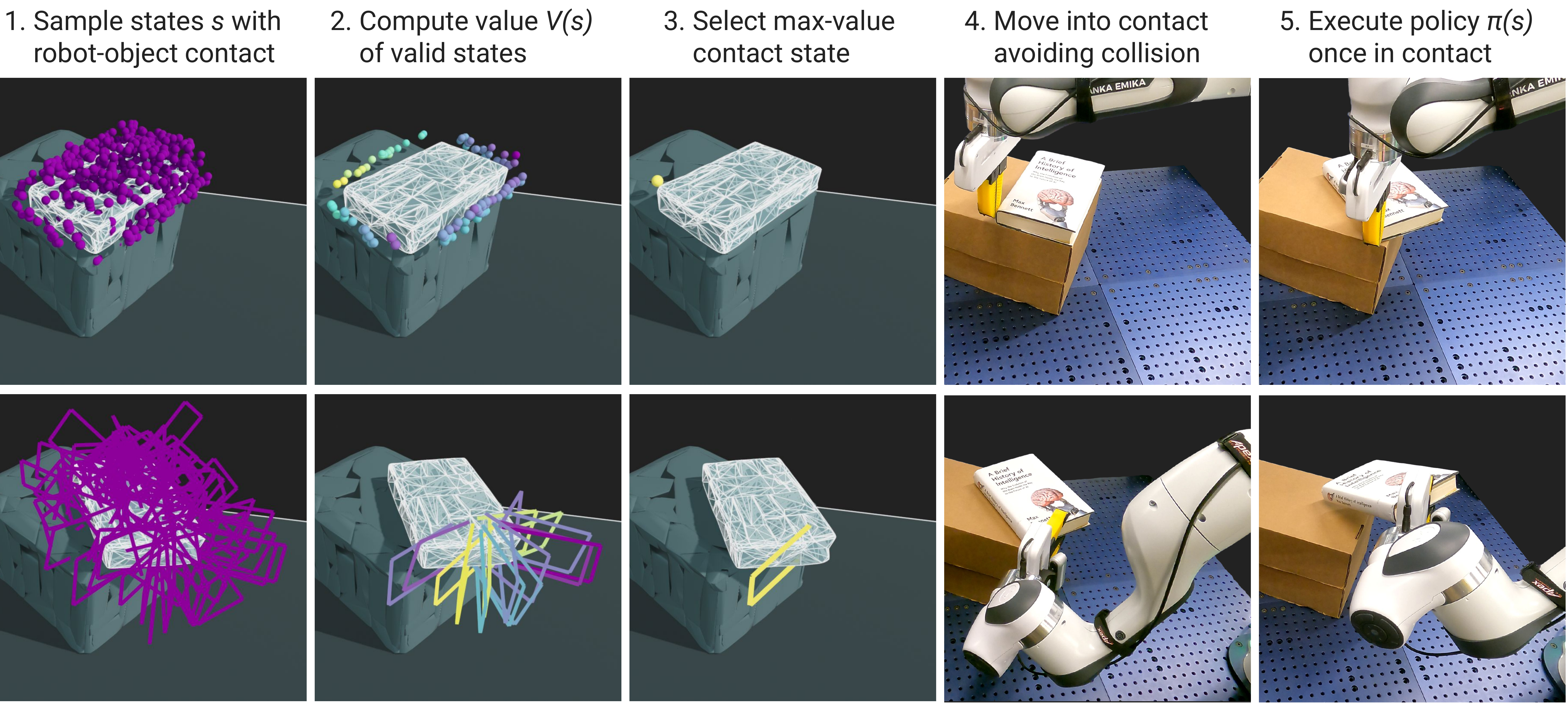}
    \caption{An overview of our contact action space on a multi-stage nudge-and-grasp task. (1) First we uniformly sample contact locations from the surface of the generated mesh (a subset of contacts is shown here for visual clarity). Each point shows a sampled gripper pose for nudging (top row) or grasping (bottom row). (2) Then invalid contact states are eliminated, e.g. those colliding with the rest of the object, and the learned value function from RL is used to compute the value of each valid state. Purple states have lower value, and yellow states have higher value. (3) At execution, the max-value state is chosen. (4) The robot moves into contact using motion planning, and then (5) executes policy actions. Meshes shown have been approximated for fast collision checking \cite{vhacd}. This action space enables policy learning from scratch in minutes.}
    \label{fig:contact}
\end{figure*}

\textbf{Motivation for action space}. We require an action space which enables learning from scratch in minutes, which experiments show is difficult with standard end-effector control (Section \ref{ss:ablations}). As we have a complete generated object mesh, we propose leveraging this to accelerate exploration. The key insight is that at some point, to complete the task, there must be a state where the robot makes contact with the object for the first time. We can sample many contact states $s$, use the learned value function from RL to compute $V(s)$, and move the robot directly into a high-value contact state. Intuitively, this accelerates exploration because we avoid spending simulator time on free-space motion. The robot immediately starts interacting and collecting reward signal.

\textbf{Action space algorithm}. An overview is shown in Figure \ref{fig:contact}. We assume that the gripper can either be open (for grasping) or closed (for pushing). The allowed contact types vary by experiment (Section \ref{ss:exp-setup}). We sample many gripper poses in parallel. First, we uniformly sample contact points on the surface of the generated mesh, and randomly sample gripper orientations guided by the mesh surface normals. Then we eliminate contact poses which are physically invalid, using cuRobo \cite{curobo} for GPU-accelerated collision checking between the gripper and object/background. For the remaining poses, we construct an ``imaginary'' state vector $s$ which matches the current state of the scene, except that the gripper is at the sampled contact pose. We then compute the value of each contact state using the learned value function $V(s)$. At test time, we move the robot into the highest-value contact pose with motion planning, whereas during training we select from $N$ sampled contact states with a probability $p(s)$ using softmax to allow for exploration, as shown in \eqref{eq:contact-probs}.

\begin{equation}\label{eq:contact-probs}
    p(s) = (1 - K) \times \frac{e^{V(s)/T}}{\sum_{i=1}^N e^{V(s_i)/T}} + K \times \frac{1}{N}
\end{equation}

The temperature $T$ and uniform sampling probability $K$ balance exploitation with exploration. We also use ``value function warmup'': we train with a high $K$ value (mostly uniform contact sampling) for several episodes, to be less vulnerable to unfortunate value function initialization. Once in contact, the robot executes actions from its learned policy $\pi(s)$. The output actions are 6D relative end-effector movements. To further accelerate exploration, the policy learns residual actions \cite{residual-learning} which are added on top of the down-scaled goal flow vector. For multi-stage tasks, after the robot has finished executing its policy actions from the first contact, it samples contact states again, moves into contact elsewhere, executes policy actions, and repeats until timeout.

\textbf{Real-robot execution}. The trained policy from simulation is deployed zero-shot on the real robot. We select the highest-value sampled contact pose which has a collision-free inverse kinematics solution. While the learned value function from PPO is typically discarded after training, we show how it can be reused to evaluate sampled contact states during deployment, a key insight for rapid learning. The robot then computes a motion plan using cuRobo \cite{curobo} to move into that contact pose while avoiding collisions with the object and the environment, and then executes actions from the policy.

\begin{figure}[ht]
    \centering
    \includegraphics[width=1.0\linewidth]{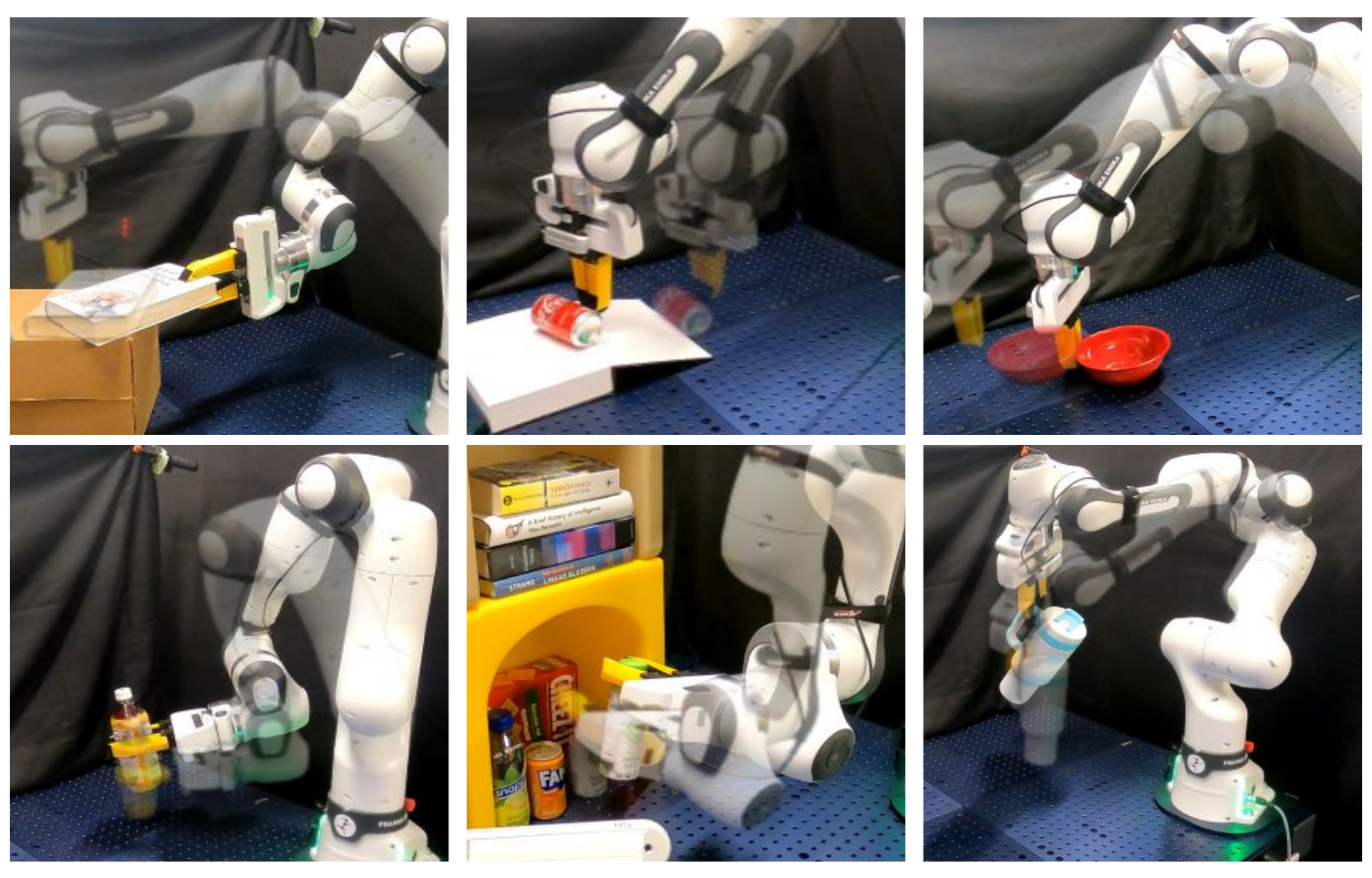}
    \caption{The six tasks studied for the end-to-end real world evaluation. Top: \textit{MultiStageBook}, \textit{PushUpRamp}, \textit{SlideBowl}. Bottom: \textit{NovelPoses}, \textit{PullFromShelf}, and \textit{GraspHandle}.}
    \label{fig:all-tasks}
    \vspace{-0.2cm}
\end{figure}

\section{Experimental Evaluation}

\subsection{Research Questions}

We investigate the following questions: \textbf{(1)} Is it possible to learn manipulation tasks from scratch in minutes, without human demonstrations or policy pre-training, given the assumptions of our problem setting? \textbf{(2)} For sim2real transfer, how important is it to have a complete mesh (generated using a 3D prior), compared to a partial mesh from a depth image? \textbf{(3)} How does mesh generation compare to mesh retrieval for real-to-sim-to-real? \textbf{(4)} How well does our method perform under challenging viewpoints, for example when task-critical parts of an object are barely visible, or not visible at all? \textbf{(5)} Which design decisions contribute the most to accelerating training (Section \ref{ss:ablations})? \textbf{(6)} Can our framework be extended to articulated objects? \textbf{(7)} How does mesh generation compare with using a ground-truth scanned mesh (Section \ref{ss:artic-exp})?

\subsection{Experiment Setup}\label{ss:exp-setup}

\noindent \textbf{Hardware}. We use a 7-DoF Franka Research 3 arm with a compliant gripper. A wrist camera (RealSense D435) is used to capture an RGB-D object image from a specified viewpoint for input to the image-to-3D pipeline, and to reconstruct the background. An external RGB-D camera (RealSense D455) tracks the object with Foundation Pose during execution using one A6000 GPU. Another A6000 is used for policy training with 256 parallel agents in simulation.

\noindent \textbf{Evaluation tasks}. We conduct a thorough end-to-end evaluation across a set of six real-world tasks (Figure \ref{fig:all-tasks}) listed below. Despite the challenging conditions of our problem setting (learning from scratch in minutes with zero demonstrations), our method enables the robot to perform a range of everyday tasks. The tasks are similar in difficulty to real-robot tasks from related work in real2sim \cite{acdc,video2policy}.

1. \textit{MultiStageBook}: the robot must learn to nudge the book to the side of the box to expose valid grasps (but not so far that it falls off), and then perform a 6-DoF grasp while avoiding collisions. This task is challenging because it requires multi-stage, mixed-contact manipulation, interacting with both sides of the object, using the same $\pi(s)$ and $V(s)$. Here the action horizon is two contact actions (where only push contacts can be sampled for the first action, and only grasps for the second). In other tasks it is one contact action.\\
2. \textit{PushUpRamp}: the robot must push or roll a Coca-Cola can up a ramp, to the top. This requires precise balance. The geometry of the unseen surfaces of the can must be correctly generated for accurate rolling dynamics in simulation.\\
3. \textit{SlideBowl}: the robot must slide the bowl along the table towards the robot, by pushing from the far side of the bowl: this unseen side of the object must be generated accurately.\\
4. \textit{NovelPoses}: the robot must grasp and lift a tea bottle. During training, the object is initialized randomly within a 20x20 cm area with a random z-axis orientation. At test time, the robot must grasp the object from a random novel pose within that area, testing if the learned policy can generalize.\\
5. \textit{PullFromShelf}: the robot must pull a bottle from a cluttered shelf where most grasps would be invalid due to collisions. Here, scanning the object would be impossible due to occlusions, whereas our method only needs \textit{one} clear view of the object to generate a complete mesh.\\
6. \textit{GraspHandle}: to grasp the tumbler, the robot must generate a complete mesh from a challenging input view, where the handle is only partially visible due to the object's self-occlusion, but it is critical for completing the task.

\textbf{Task success} is defined using a goal region. For \textit{MultiStageBook}, the book's center of mass must be raised: just grasping counts as failure. For examples of success in each task, please see the videos at: \href{https://zerobot-rl.github.io}{zerobot-rl.github.io}

\textbf{Baselines}. \textit{ZB-NoPrior} is an ablation of our method which constructs a partial object mesh using Poisson Surface Reconstruction~\cite{poisson-surface-recon} from the single-view point cloud, without any 3D prior. We also compare with a real2sim framework called ACDC \cite{acdc}, which shows impressive scene-level reconstruction. The \textit{ACDC-NN} (nearest neighbor) baseline retrieves the most similar object mesh from a database based on the input image, whereas ZeroBot generates a mesh. For a fair comparison between these real2sim methods, they all use our RL training framework after building the simulation.

\begin{figure*}[ht]
    \centering
    \includegraphics[width=0.87\linewidth]{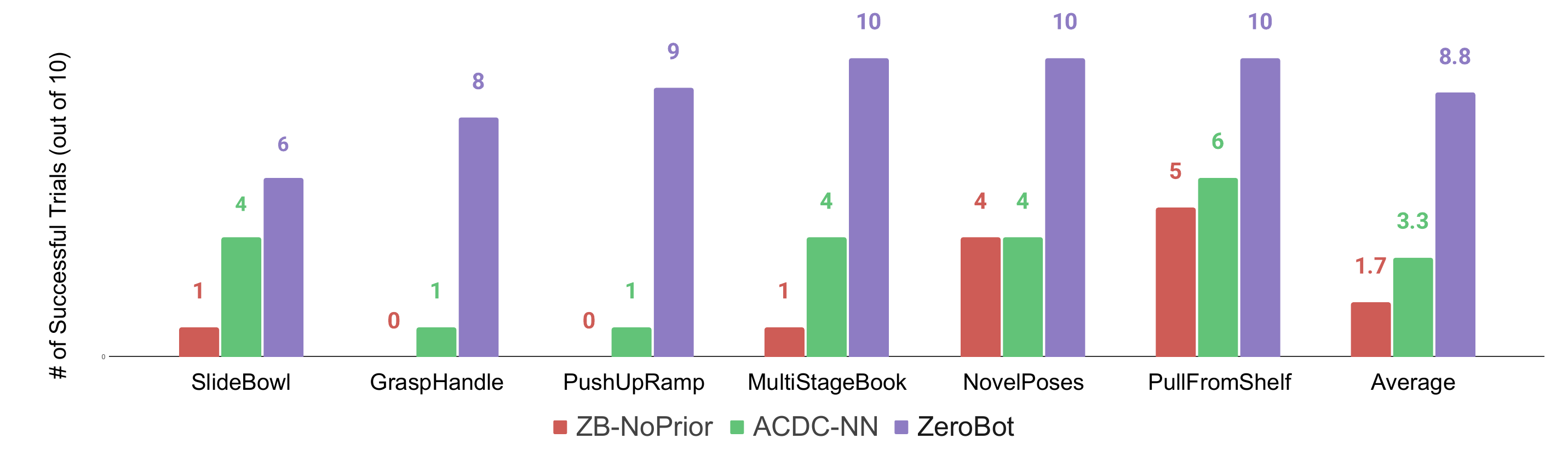}
    \caption{Success rates on six real-robot evaluation tasks, comparing our method ZeroBot against the baseline methods \textit{ZB-NoPrior} and \textit{ACDC-NN} \cite{acdc}. We conduct 10 end-to-end repeats for each method and for each task, generating a new mesh and training a policy from scratch for each repeat (180 in total). ZeroBot's generated meshes lead to higher success rates.}
    \label{fig:quant-results}
\end{figure*}

\begin{figure*}[ht]
    \centering
    \includegraphics[width=0.84\linewidth]{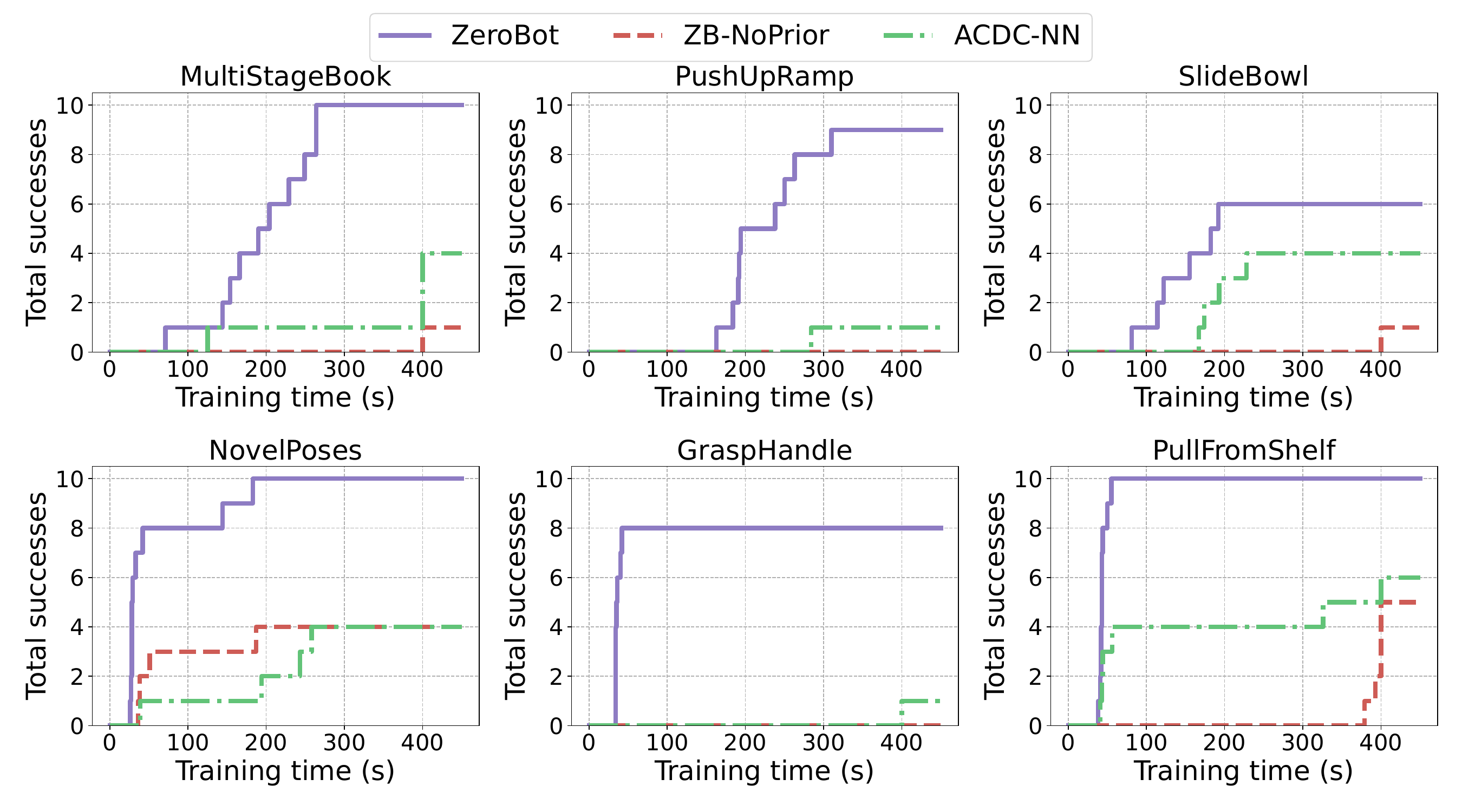}
    \vspace{-0.1cm}
    \caption{Step curves showing cumulative real world successes (out of 10) against wall-clock training time. Each point on the curve shows how many training runs finished by that time \textit{and} led to a real-world success. Training finishes when a threshold success rate has been reached in simulation, or when the time limit is reached at 400 seconds. This shows that our method can learn to manipulate a novel object for useful tasks in minutes, with no demonstrations or policy pre-training.}
    \label{fig:real_curves}
    \vspace{-0.2cm}
\end{figure*}

\subsection{Real2Sim2Real Evaluation Results}

The quantitative results of this real-world evaluation are in Figure \ref{fig:quant-results}. Below we provide analysis and highlight each of the main conclusions, answering our research questions.

\textbf{Conclusion 1:} \textit{A complete object mesh is important for accurate simulation dynamics, robust tracking, and sim2real transfer.} On all tasks, our ZeroBot method out-performs the \textit{ZB-NoPrior} baseline which does not use a 3D prior. In \textit{PushUpRamp}, the incomplete mesh makes it difficult for the baseline to succeed in simulation, whereas our framework generates a round, smooth, and symmetric mesh. In the GraspHandle and MultiStageBook tasks, the baseline often collides when attempting to transfer a grasp that succeeds in simulation with the partial mesh onto the real object.

\begin{figure}[htb]
    \centering
    \includegraphics[width=\linewidth]{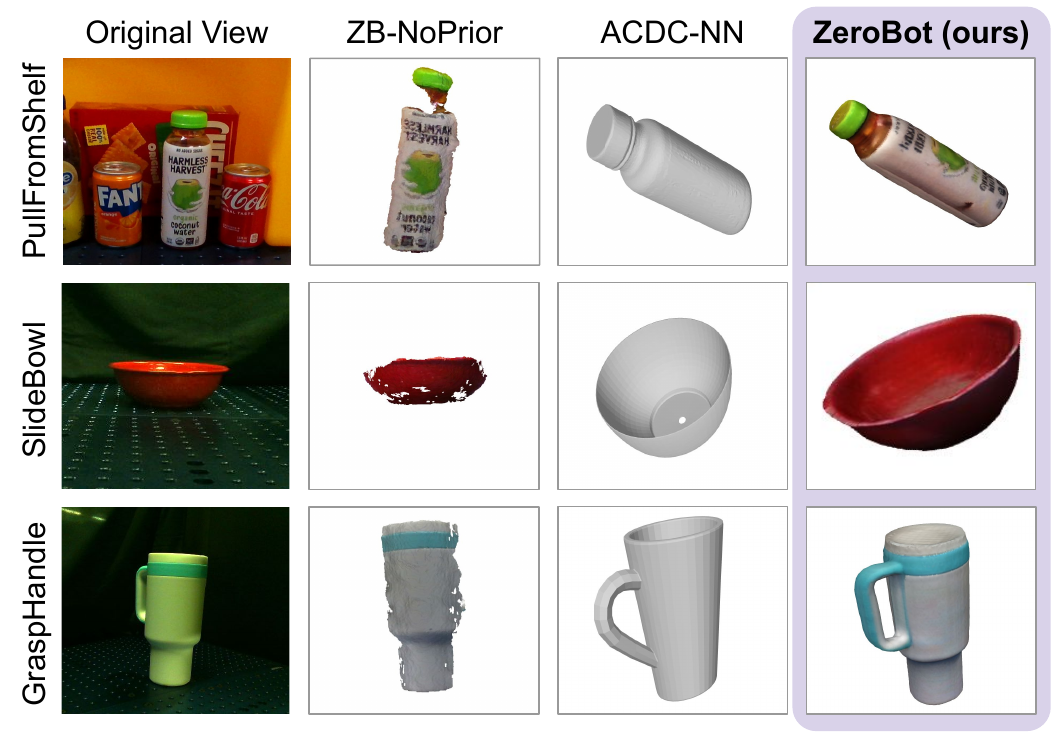}
    \caption{Qualitative results comparing mesh generation methods. ZeroBot generates a mesh which is complete and more accurate than the baselines, improving sim2real success rates.}
    \label{fig:meshes}
    \vspace{-0.3cm}
\end{figure}

\textbf{Conclusion 2}: \textit{Generating a mesh ad-hoc enables greater precision than retrieving the nearest neighbor from a model dataset}. While the \textit{ACDC-NN} \cite{acdc} baseline performs reasonably on common objects such as bottles, our method achieves higher success rates on less common objects such as the tumbler. Grasps which may be valid on the nearest neighbor may not be valid on the real object (see Figure \ref{fig:meshes}). Additionally, our method performs better in higher-precision tasks e.g. \textit{PushUpRamp}. Our results show that an image-to-3D model can generate sufficiently accurate meshes for sim2real transfer, enabling rapid real2sim for novel objects.

\textbf{Conclusion 3:} \textit{Accurate mesh generation is possible even from challenging views where task-critical geometry is unobserved}. The results from \textit{SlideBowl} and \textit{GraspHandle} show that using image-to-3D models, our framework can generate accurate geometry which enables sim2real transfer even when the robot interacts with a part of the object which it has never directly observed, such as the far side of the bowl, or the partially-occluded tumbler handle.

\textbf{Conclusion 4:} \textit{ZeroBot can learn to manipulate from scratch in minutes}, within our set of assumptions. Figure \ref{fig:real_curves} shows training times required for real-world successes. By using visual foundation models for real2sim, and our efficient action space, ZeroBot can rapidly learn policies without any human demonstrations or policy pre-training.

\subsection{Training Speed Comparison}\label{ss:ablations}

\begin{figure}[hbt]
    \centering
    \includegraphics[width=\linewidth]{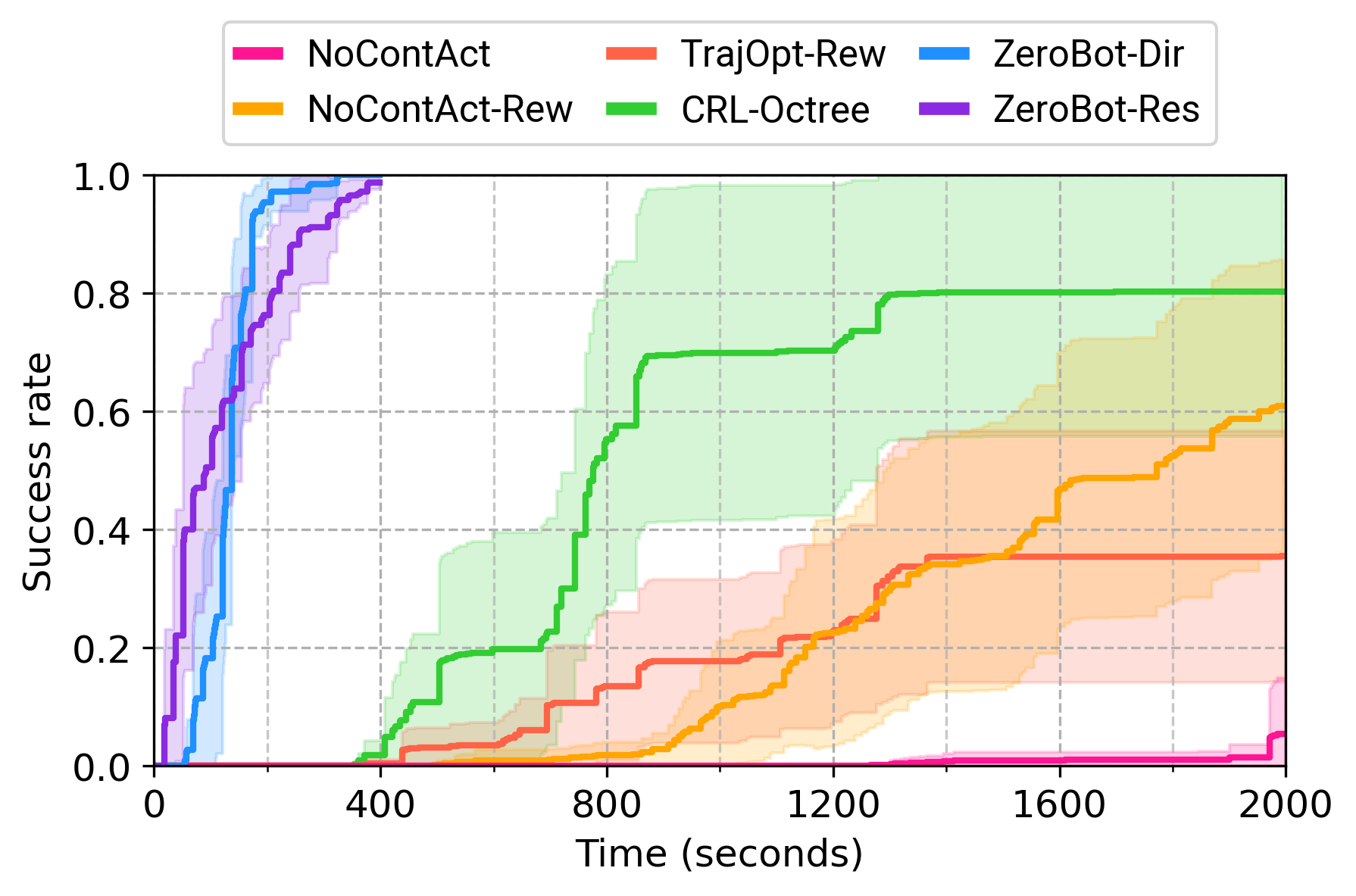}
    \caption{Training times (wall-clock) for each method. Each checkpoint is evaluated in simulation across 256 parallel environments for 10 episodes. Each point on the curve shows the maximum success rate achieved by that run so far. This is averaged across 10 training runs, and the 95\% confidence interval for the mean is plotted. Results show that ZeroBot's action space significantly accelerates training.}
    \label{fig:training_speed_curves}
\end{figure}

Figure \ref{fig:training_speed_curves} shows how design choices affect training speed. We focus on the SlideBowl task, chosen because all baselines can achieve some task progress. The \textit{NoContAct} ablation does not use our action space (only a Cartesian policy) and barely learns after 30 minutes. \textit{NoContAct-Rew} adds an auxiliary reward encouraging gripper-object contact, which is used in Gen2Sim \cite{gen2sim}. \textit{TrajOpt-Rew} uses CEM with covariance clamping \cite{drop} and executes open-loop, but its performance plateaus. ZeroBot learns much faster, outputs a closed-loop policy, and does not need any auxiliary rewards. Instead of sampling contact poses from the object surface, \textit{CRL-Octree} uses Coarse-to-fine Reinforcement Learning \cite{crl} to iteratively ``zoom in'' to high-value regions of the scene using an octree. ZeroBot still trains faster, showing that leveraging the generated geometry as part of the action space accelerates exploration. We also compare two variations of our method: once in contact, \textit{ZeroBot-Res} (our main method) predicts residual actions on top of the goal flow vector, whereas \textit{ZeroBot-Dir} directly predicts actions. Residual learning leads to slightly faster exploration early in training, though both variants significantly outperform the baselines. This shows that our contact action space is the key factor which enables learning from scratch in minutes.

\subsection{Extending our Framework to Articulated Objects}\label{ss:artic-exp}

\begin{figure}[hbt]
    \centering
    \includegraphics[width=1.0\linewidth]{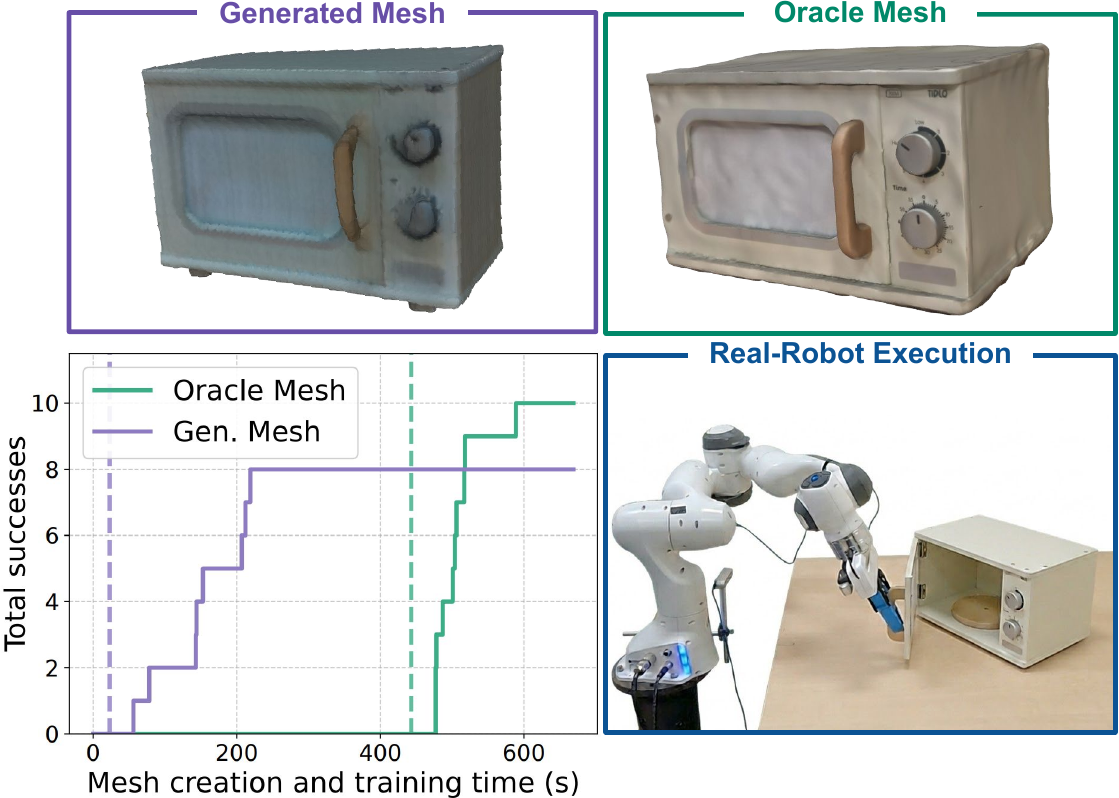}
    \caption{\textbf{Top:} generated mesh and oracle mesh from manual scanning for the articulated object. \textbf{Bottom:} real-to-sim-to-real results, showing cumulative successes against time (left), where the dashed lines indicate the end of mesh creation, and the real robot successfully executing the policy (right).}
    \label{fig:artic_combined}
\end{figure}

We demonstrate how our framework can be extended to articulated tasks. In this experiment, the robot needs to grasp and open the microwave door.  As before, InstantMesh is used to generate the mesh for the whole object from a single view. We assume that the articulation parameters (e.g. joint limits, door segmentation) are given (as in \cite{rialto,gen2sim}). In future work, these can be inferred automatically \cite{urdformer}. We track the pose of the articulated part (i.e. the door) with Foundation Pose, and use this as input to the policy. We compare single-view mesh generation with a real2sim method used in RialTo \cite{rialto}: the mesh is reconstructed using Polycam photogrammetry from a manual scan of the object with 164 views. This oracle baseline allows us to measure the performance gap between generated meshes and ground-truth geometry. Both methods use our contact action space for policy learning, as RialTo originally requires demonstrations and takes over 2 days to train, which makes a system-level comparison infeasible in our problem setting. In Figure \ref{fig:artic_combined} we plot the time taken to reach real-robot successes, including both training time and mesh creation time (23 seconds for mesh generation, and 443 seconds for creating the oracle mesh, including an allowance of 290 seconds for manual scanning, as per the user study in RialTo \cite{rialto}). As expected, ground-truth geometry leads to higher sim2real success rates. However, single-view mesh generation still achieves 80\%, and has other advantages: it is significantly faster and more autonomous than a manual scan.

\section{Limitations and Future Work}

ZeroBot inherits the limitations of current image-to-3D models, such as precision \cite{image-3d-limits}. As these models improve, for example through multi-view input \cite{trellis} or occlusion-aware generation \cite{amodal3r}, future work can incorporate them to relax our static background assumption and enable a broader range of multi-object tasks in more dynamic scenes. In this paper, we focus on real2sim for geometry. Following RialTo \cite{rialto}, we set a constant mass across objects in simulation, and we assume that articulation parameters are known. Inferring physics \cite{asid} and articulation properties \cite{urdformer} automatically is a complementary research direction, which future work can incorporate. For faster and smoother execution, future work can use asynchronous policies, or use our real2sim framework for trajectory optimization rather than policy learning.

\section{Conclusion}
\label{s:conclusion}

We presented ZeroBot, a real2sim framework for learning a policy from scratch in minutes. By leveraging image-to-3D models, ZeroBot generates an object mesh and learns to manipulate it in simulation, requiring zero human demonstrations or policy pre-training. Despite the challenging problem setting, ZeroBot can learn a range of useful tasks. Real-robot experiments showed that generated meshes enable successful sim2real transfer, out-performing baselines using partial meshes or nearest neighbor retrieval. Ablation studies confirmed that our action space significantly accelerates training. Our work demonstrates that image-to-3D models, although imperfect, are a powerful real2sim technique with significant potential for enabling rapid, autonomous robot learning.

\section*{Acknowledgments}

The authors would like to thank Andrew Davison, Shikun Liu, Ran Gong, Karl Schmeckpeper,  Jad Abou-Chakra, Lingfeng Sun, Tushar Kusnur, Gary Lvov, Yifei Ren, Eric Dexheimer, Marwan Taher, Kamil Dreczkowski, Norman Di Palo, Vitalis Vosylius, Georgios Papagiannis, and Pietro Vitiello for helpful discussions. Ivan Kapelyukh was partly supported by Dyson Technology Ltd and EPSRC Prosperity Partnerships (EP/S036636/1).

\ifCLASSOPTIONcaptionsoff
  \newpage
\fi

\bibliographystyle{IEEEtran}
\bibliography{zerobot}

\end{document}